\documentclass{article}

\usepackage{arxiv}

\usepackage[utf8]{inputenc} 
\usepackage[T1]{fontenc}    
\usepackage{hyperref}       
\usepackage{url}            
\usepackage{booktabs}       
\usepackage{amsfonts}       
\usepackage{nicefrac}       
\usepackage{microtype}      
\usepackage{lipsum}		
\usepackage{graphicx}
\usepackage{natbib}
\usepackage{doi}

\usepackage{subcaption}
\usepackage{stfloats}  
\usepackage{amsmath,amsthm,amssymb,amsbsy,amssymb,mathstyle}
\usepackage{tikz}
\usepackage{array}

\newdimen\nodeDist
\newcolumntype{L}[1]{>{\raggedright\let\newline\\\arraybackslash\hspace{0pt}}m{#1}}
\newcolumntype{C}[1]{>{\centering\let\newline\\\arraybackslash\hspace{0pt}}m{#1}}
\newcolumntype{R}[1]{>{\raggedleft\let\newline\\\arraybackslash\hspace{0pt}}m{#1}}

\newcommand{\E}{\mathbf{E}}
\newcommand{\N}{\mathbf{N}}
\newcommand{\IG}{\mathbf{IG}}
\newcommand{\X}{\mathbf{X}}
\newcommand{\x}{\mathbf{x}}
\newcommand{\m}{\mathbf{m}}
\newcommand{\y}{\mathrm{y}}
\newcommand{\res}{\mathrm{r}}
\newcommand{\mures}{\mathrm{m}}
\newcommand{\vres}{\mathrm{v}}
\newcommand{\taures}{\mathrm{t}}
\newcommand{\bvec}{\mathrm{b}}
\newcommand{\epsy}{\epsilon}
\newcommand{\iter}{\text{iter}}
\newcommand\independent{\protect\mathpalette{\protect\independenT}{\perp}}
\def\independenT#1#2{\mathrel{\rlap{$#1#2$}\mkern2mu{#1#2}}}

\usepackage{algorithm}
\usepackage{algorithmic}
\usepackage{multirow}

\title{Stochastic Tree Ensembles for Estimating Heterogeneous Effects}

\author{ Nikolay Krantsevich \\
    School of Mathematical and Statistical Sciences \\
	Arizona State University\\
	Tempe, Arizona, USA \\
	\And
	Jingyu He \\
	Department of Management Sciences \\
	City University of Hong Kong \\
	Hong Kong SAR \\
	 \And
	P. Richard Hahn \\
    School of Mathematical and Statistical Sciences \\
	Arizona State University\\
	Tempe, Arizona, USA \\
}

\renewcommand{\shorttitle}{Stochastic Tree Ensembles for Estimating Heterogeneous Effects}

\hypersetup{
pdftitle={Stochastic Tree Ensembles for Estimating Heterogeneous Effects},
pdfauthor={Nikolay Krantsevich, Jingyu He, PR Hahn},
pdfkeywords={Machine Learning, Causal Inference, Heterogeneous effects},
}

\begin{document}
\maketitle

\begin{abstract}
  Determining subgroups that respond especially well (or poorly) to specific interventions (medical or policy) requires new supervised learning methods tailored specifically for causal inference. Bayesian Causal Forest (BCF) is a recent method that has been documented to perform well on data generating processes with strong confounding of the sort that is plausible in many applications. This paper develops a novel algorithm for fitting the BCF model, which is more efficient than the previously available Gibbs sampler. The new algorithm can be used to initialize independent chains of the existing Gibbs sampler leading to better posterior exploration and coverage of the associated interval estimates in simulation studies. The new algorithm is compared to related approaches via simulation studies as well as an empirical analysis.
\end{abstract}


\section{Background}
\label{RIC}

\subsection{Estimating heterogeneous effects}
This paper considers the use of supervised machine learning for estimating conditional average treatment effects (CATE), the treatment effect averaged across subpopulations defined in terms of measured attributes. 

Let $Y_i$ represent the scalar response variable, $Z_i$ denote a binary treatment variable, and $\x_i$ represent a length $d$ row vector of observed control variables for observation $i$. Let $Y$ and $Z$ be length $n$ column vectors comprising variables $Y_i$ and $Z_i$ respectively; let $\X$ denote the $n \times d$ matrix of control variables.   We will use lower case Roman letters, such as $y$ and $z$, to denote the values assumed by variables. Our data will consist of $n$ independent observations $(Y_i, Z_i, \x_i)$.

Following the potential outcomes framework \citep{imbens2015causal}, let $Y_i(1)$ and $Y_i(0)$ represent the outcomes under treatment and control respectively; each observed response may be expressed as $Y_i = Z_i Y_i(1) + (1 - Z_i) Y_i(0)$. 

Throughout, we assume the following standard conditions licensing regression estimates of treatment effects:
\begin{enumerate}
  \item \textbf{SUTVA (Stable Unit Treatment Value Assumption)} implies that no treatment assignment to a particular individual should affect the observed outcomes on other individuals and that there is no variation in treatment.
  \item \textbf{Strong ignorability assumption} implies that, first, there are no unmeasured confounders:
\begin{equation}\label{ignore1}
 Y_i(0), Y_i(1) \independent Z_i \mid \X_i,
\end{equation}
and, second, that every individual has a non-zero probability of being assigned to treatment:
 \begin{equation}\label{ignore2}
 0 < \mbox{Pr}(Z_i = 1 \mid \x_i) < 1.
\end{equation}
\end{enumerate}
Under these assumptions, the conditional average treatment effect of units with covariates $\x$ may be estimated as the differences of two identified conditional expectations:
\begin{equation}\label{TE}
  \tau(\x) := \E(Y \mid \x, Z = 1) -  \E(Y \mid \x, Z = 0).
\end{equation}
Further assuming a mean-zero additive error, 
\begin{equation}\label{response}
  Y_i = f(\x_i, Z_i) + \epsy_i,\quad \epsy_i\sim \N(0,\sigma^2),
\end{equation}
it follows that $\E(Y_i \mid \x_i, Z_i=z_i) = f(\x_i, z_i)$ and 
\begin{equation}\label{TE2}
  \tau(\x_i) := f(\x_i, 1) - f(\x_i, 0).
\end{equation}

Here, these conditional expectations will be estimated using a Bayesian tree ensemble method \cite{he2019xbart} related to a well-known method called Bayesian additive regression trees, or BART  \citep{chipman2010bart}.



\subsection{BART model and prior}\label{BART}
\begin{figure}[b]
  \centering
  \begin{subfigure}{.5\columnwidth}
    \centering
    \begin{tikzpicture}[
        scale=0.7,
        node/.style={%
            draw,
            rectangle,
          },
        node2/.style={%
            draw,
            circle,
          },
      ]

      \node [node] (A) {$x_1<0.8$};
      \path (A) ++(-135:\nodeDist) node [node2] (B) {$m_{l1}$};
      \path (A) ++(-45:\nodeDist) node [node] (C) {$x_2<0.4$};
      \path (C) ++(-135:\nodeDist) node [node2] (D) {$m_{l2}$};
      \path (C) ++(-45:\nodeDist) node [node2] (E) {$m_{l3}$};

      \draw (A) -- (B) node [left,pos=0.25] {no}(A);
      \draw (A) -- (C) node [right,pos=0.25] {yes}(A);
      \draw (C) -- (D) node [left,pos=0.25] {no}(A);
      \draw (C) -- (E) node [right,pos=0.25] {yes}(A);
    \end{tikzpicture}
  \end{subfigure}%
  \begin{subfigure}{.5\columnwidth}
    \centering
    \begin{tikzpicture}[scale=2.5]
      \draw [thick, -] (0,1) -- (0,0) -- (1,0) -- (1,1)--(0,1);
      \draw [thin, -] (0.8, 1) -- (0.8, 0);
      \draw [thin, -] (0.0, 0.4) -- (0.8, 0.4);
      \node at (-0.1,0.4) {0.4};
      \node at (0.8,-0.1) {0.8};
      \node at (0.5,-0.2) {$x_1$};
      \node at (-0.3,0.5) {$x_2$};
      \node at (0.9,0.5) {$m_{l1}$};
      \node at (0.4,0.7) {$m_{l2}$};
      \node at (0.4,0.2) {$m_{l3}$};
    \end{tikzpicture}
  \end{subfigure}
  \caption{(Left) An example binary tree, with internal nodes labelled by their splitting rules and terminal nodes labelled with the corresponding parameters $m_{lb}$. (Right) The corresponding partition of the sample space and the step function. Here $\m_l = (m_{l1}, m_{l2}, m_{l3})$.}
  \label{fig:treestep}
\end{figure}
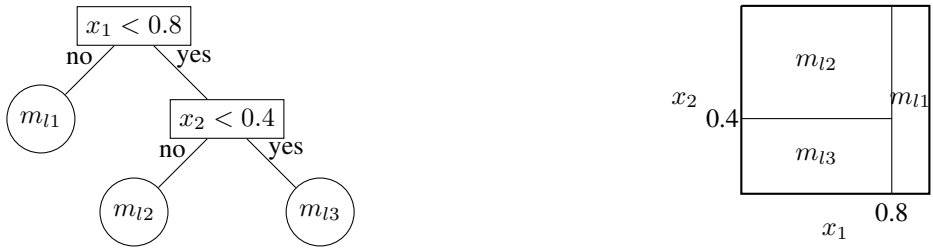

BART represents the outcome of interest as a sum of an unknown function $f(\cdot)$ and an error term,
\begin{equation}\label{BARTrepresentation}
  Y_i = f(\x_i) + \epsy_i, \quad \epsy_i\sim \N(0,\sigma^2)
\end{equation}
The mean function $f(\x)$ is represented as a sum of many piecewise constant binary regression trees
\begin{equation}\label{forest}
  f(\x)=\sum_{l=1}^L g_l(\x; T_l, \m_l)
\end{equation}
where $T_l$ denotes a regression tree, which represents a partition of the covariate space (say $\mathcal{A}_1,\dots,\mathcal{A}_{B(l)}$) and consists of a set of internal decision nodes and a set of terminal nodes (or leaves) which correspond to each element of the partition. Each element of the partition $\mathcal{A}_b$ is associated a leaf parameter value, $m_{lb}$, and $\m_l = (m_{l1}, \cdots, m_{lB(l)})$ denotes a vector corresponding to all leaf parameters of the $l$-th tree, $T_l$. The piecewise constant function comprising the partition and the leaf parameters is defined as $g_l(\x) = m_{lb}\ \text{if}\ \x\in \mathcal{A}_b$; see Figure \ref{fig:treestep} for demonstration.

Within each leaf, the mean parameters are given independent normal priors, $m_{lb} \sim \N(0, \nu)$. The prior over trees $p(T_l)$ is specified by the probability of a node having children at depth $d$ \cite{chipman1998bayesian} as
\begin{equation}\label{eq:BARTprior}
  \alpha (1+d)^{-\beta}, \qquad
  \alpha \in (0,1), \beta \in [0, \infty)
\end{equation}

BART explores the posterior of the trees by random walk Metropolis-Hastings Markov chain Monte Carlo (MCMC) algorithm, which can be slow to converge and limits the broader adoption of BART for large datasets.


\subsection{XBART} \label{XBART}

XBART, short for Accelerated Bayesian Additive Regression Trees \cite{he2020stochastic}, was introduced to improve the fitting time of BART-like models. XBART blends regularization and stochastic search strategies from Bayesian modeling with computationally efficient techniques from recursive partitioning approaches to tree-fitting. XBART fits the same sum-of-trees ensemble model as BART but regrows each tree  \textit{recursively} at each iteration according to a stochastic process inspired by Bayesian updating.

We review the stochastic tree-growing approach of XBART (Algorithm \ref{alg:GFR}).  Let $\mathcal{C}$ denote a matrix of cutpoint candidates, with each element $c_{jk}$ where $j = 1,\cdots,p$ indexes a variable and $k$ indexes a candidate cutpoint. Assume the leaf parameter $m$ has prior $N(0, \nu)$. At each node, the probability of splitting at cutpoint $c_{jk}$ is proportional to
\begin{equation}\label{eqn:split_gaussian}
  {\small
    \begin{aligned}
      L(c_{jk}) & \propto \exp \left\{\frac{1}{2}\left[ \log{ \left( \frac{\sigma^2}{\sigma^2 + \nu n_{jk}^l} \right)}  + \right.\right. \\
     & \left.\left.+  \frac{\nu}{\sigma^2(\sigma^2  + \nu n_{jk}^l)} \left(s_{jk}^l\right)^2 +   \log{ \left( \frac{\sigma^2}{\sigma^2 + \nu n_{jk}^r} \right)}  \right.\right. \\
                & \left.\left. + \frac{\nu}{\sigma^2(\sigma^2 + \nu n_{jk}^r)} \left(s_{jk}^r\right)^2 \right]\right\},
    \end{aligned}
  }
\end{equation}
where $\sigma^2$ is the residual variance as in equation (\ref{BARTrepresentation}), $n_{jk}^l$ and $ n_{jk}^r$ correspond to the number of data observations on left or right child node if split at the splitting rule $c_{jk}$, and $s_{jk}^l$ and $s_{jk}^r$ are the corresponding sufficient statistics for the children nodes:
\begin{equation}\label{eqn:suffstats}
  {\small
    \begin{gathered}
      s_{jk}^l = \sum_{x_i \in \mathcal{A}^{\text{left}}_{jk}} y_i, \quad s_{jk}^r = \sum_{x_i \in \mathcal{A}^{\text{right}}_{jk}} y_i\\
      s_{\text{all}} = s_{jk}^l + s_{jk}^r = \sum_{i = 1}^n y_i,
    \end{gathered}
  }
\end{equation}
where $n = n_{jk}^l + n_{jk}^r$ is number of observations in the current node. Similarly, the probability of not splitting anywhere is proportional to
\begin{equation}\label{eqn:split_gaussian2}
  {\small
    \begin{aligned}
      L(\emptyset) & \propto |\mathcal{C}| \left(\frac{(1+d)^\beta}{\alpha} - 1\right) \times                                                                                          \\
                   & \exp\left\{\frac{1}{2}\left[\log{ \left( \frac{\sigma^2}{\sigma^2 + \nu n} \right)}  + \frac{\nu}{\sigma^2(\sigma^2 + \nu n)} s_{\text{all}}^2\right]\right\}.
    \end{aligned}
  }
\end{equation}
where $|\mathcal{C}|$ is the total number of candidate splitting rules, $d$ is the depth of the current node in the tree. The tree is fitted recursively where at each node, a cutpoint (or the stop-splitting option) is randomly drawn from a multinomial distribution using probabilities of (\ref{eqn:split_gaussian}) and (\ref{eqn:split_gaussian2}). If stop-splitting is sampled, or other pre-set stopping conditions are satisfied, the current node becomes a terminal (leaf) node, and its associated leaf parameter $m$ is updated by conjugate Gaussian sampling. 
To form an ensemble of trees, XBART uses a similar strategy as Bayesian backfitting, residualizes the data with respect to the partial fit corresponding to the forest. Specifically, the $h$-th tree is grown to fit the partial residual of all other trees: $\y - \sum_{l\neq h}g_l(\x; T_l, \m_l)$.

\citet{he2020stochastic} details how these strategies contribute to the improved efficiency of XBART over BART as well as improved posterior coverage of interval estimates obtained by initializing multiple Markov chains at XBART estimates. Here, we adapt the XBART approach to the BCF model and demonstrate comparable performance gains in the heterogeneous treatment effect setting.

\begin{algorithm}[t!]
  \small
  \caption{Grow From Root (GFR)}\label{alg:GFR}
  \begin{algorithmic}[1]
    \STATE {\bfseries Input:} $\text{GFR}(\y, \X,\sigma, d, T, {\tt node})$. \\
    \STATE  Calculate full sufficient statistics $s_{\text{all}}$ by (\ref{eqn:suffstats}).
    \FOR{$c_{jk} \in \mathcal{C}$, partition data to left and right sides}
    \STATE Calculate $s_{jk}^l$ and $s_{jk}^r$ by equation (\ref{eqn:suffstats}).
    \STATE Calculate $L(c_{jk})$ by equation (\ref{eqn:split_gaussian}).
    \ENDFOR
    \STATE Calculate probability of no-split $L(\emptyset)$ by equation (\ref{eqn:split_gaussian2}).
    \STATE Draw a cutpoint or no-split using probability $L(c_{jk})$ and $L(\emptyset)$.
    \IF{no-split is chosen or stop conditions are met}
    \STATE Update leaf parameter $m_{{\tt node}}$.
    \STATE \textbf{return}.
    \ELSE
    \STATE Create two new nodes as children of ${\tt node}$, denoted ${\tt left\_node}$ and ${\tt right\_node}$.
    \STATE Sift the data into ${\tt left\_node}$ and ${\tt right\_node}$.
    \STATE  $\text{GFR}(\y_{\mbox{left}}, \X_{\mbox{left}}, \sigma, d+1, T, {\tt left\_node})$
    \STATE  $\text{GFR}(\y_{\mbox{right}}, \X_{\mbox{right}},\sigma, d+1,T, {\tt right\_node})$
    \ENDIF
    \STATE {\bfseries Output:} The grown tree $T$, including the vector of sampled leaf parameters, $\m$.
  \end{algorithmic}
\end{algorithm}

\subsection{Bayesian Causal Forest}\label{BCF}

Via simulation studies, \citet{hahn2020bayesian} demonstrate the inability of BART to handle confounding for certain simple data generative processes (DGPs). They refine the BART model to overcome this limitation with several modifications. First, rather than representing $f(\x, z)$ as a single BART model (as in \citet{hill2011bayesian}), they propose using the representation
\begin{equation}
  f(\x_i, z_i) = \mu(\x_i) + \tau(\x_i) z_i,\label{eq:genmodel}
\end{equation}
where $\mu$ and $\tau$ are prognostic and treatment functions respectively; both are given independent BART priors, permitting control and treatment effects to be regularized independently. Second, they propose including an estimate of the propensity score $\widehat{\pi}_i = P(Z_i = 1 \mid \x_i)$ as a crucial additional feature to combat unintended bias of treatment effects due to the regularization of $\mu$. See \citet{hahn2020bayesian} for more details on this phenomenon, which the authors refer to as {\em regularization induced confounding} (RIC).

Finally, \citet{hahn2020bayesian} observe that Bayesian treatment effect estimation is not invariant with respect to treatment encoding -- choosing different pairs of values as treatment indicators for treated and control groups implies different priors, which lead to different treatment effect estimates. By adding scaling factors $b_0$ and $b_1$ as parameters in the model, the priors are made invariant to which group is designated as the treated group. An additional scaling factor, $a$, is added to enhance the learning of the prognostic term.
Putting these modifications together, the Bayesian causal forest (BCF) model is
\begin{equation}\label{eq:bcfmodel}
  \begin{gathered}
    y_i = a \mu(\x_i, \widehat{\pi}_i) +  b_{z_i} \tilde\tau(\x_i) + \epsilon_i,\quad \epsilon_i\sim \N(0, \sigma^2) \\
        a \sim \N(0, 1), \quad b_0, b_1 \sim \N(0, 1/2)
  \end{gathered}
\end{equation}
According to this parametrization, treatment effects are given by $\tau(\x_i) = (b_1 - b_0)\tilde\tau(\x_i)$.


The Bayesian Causal Forest model has been documented to perform well in a number of separate, rigorous simulation studies \citep{hahn2019atlantic,dorie2018automated, wendling2018comparing}. It was recently used to estimate CATEs in the high-profile Growth Mindset intervention \cite{Yeager2019} as well as other applied work \citep{ghosh2020cpc, king2019apnoe, bail2019twitter, bryan2019replicator}.

Computationally, BCF is built upon the same random walk Metropolis-Hastings algorithm that underpins BART. As such, it suffers from the same slow fitting time on large data sets and the same slow posterior exploration. The next section seeks to address these limitations by applying the computational strategies of XBART to a BCF model.


\section{XBCF}

\subsection{The model} \label{xbcf_model}
The XBCF model differs in one substantive respect from the model presented in \citet{hahn2020bayesian}:
%
The error standard deviations $\sigma_0$ and $\sigma_1$ are allowed to differ between the control and treatment groups, respectively, whereas the original BCF model had a common shared residual standard deviation. Thus, the XBCF model is
\begin{equation} \label{eq:xbcfmodel}
\begin{gathered}
  y_i = a \mu(\x_i, \widehat{\pi}_i) +  b_{z_i} \tilde\tau(\x_i) + \epsilon_i, \;\; \epsilon_i\sim \N(0, \sigma_{z_i}^2), \\
  a \sim \N(0, 1),\;\; b_0, b_1 \sim \N(0, 1/2),
\end{gathered}
\end{equation}
or, in more detail, as
\begin{equation*}
    {\small
     \begin{gathered}
      y_i = a \sum_{l=1}^L u_l(\x_i,\widehat{\pi}_i; T_l, \m_l^{T})
      + b_{z_i} \sum_{k=1}^K v_k(\x_i; S_k, \m_k^{S}) + \epsilon_i,
       \end{gathered}
   }
\end{equation*}
where $L, K$ represent the number of trees, $T_l, S_k$ represent individual trees, $\m_l^{T}, \m_k^{S}$ denote vectors of scalar means associated with the leafs nodes of $T_l$ and $S_k$ respectively. We will reference the forests of trees as $T = \{T_l, \m_l^{T}\}_{l=1}^L$ and $S = \{S_k, \m_k^S\}_{k=1}^K$ for prognostic and treatment terms, respectively. Following BCF, we include a column vector of (estimated) propensity scores $\widehat{\pi}$ as an additional covariate for the prognostic term. 

\subsection{Modeling fitting procedure} \label{xbcf_procedure}
The XBCF fitting algorithm uses a similar ``backfitting'' strategy as BART and XBART, iterating tree-by-tree through two forests (corresponding to the prognostic and treatment terms) rather than just one. The tree and parameter updates at each iteration are based on the following ``residuals'':
\begin{equation}\label{eqn:residual}
  {\small
    \begin{aligned}
      \text{Prognostic residual: }&\vres    \equiv \y - a  \sum_{l = 1}^L u(\X,\hat{\pi};T_{l},\m_{l}^T) ,                                                   \\
      \text{Treatment residual: } &\taures  \equiv \y - \bvec \cdot \sum_{k = 1}^K v(\X;S_{k},\m_{k}^S) ,                                                         \\
      \text{Total residual: }&\res     \equiv \y - a \sum_{l = 1}^L u(\X,\hat{\pi};T_{l},\m_{l}^T)\\
      &\quad \quad  - \bvec \cdot \sum_{k = 1}^K v(\X;S_{k},\m_{k}^S).
    \end{aligned}
}
\end{equation}
where $\bvec$ is a length $n$ vector with $i$-th component equal to $b_{z_i}$, and `$\cdot$' denotes element-wise multiplication. 
The update steps for trees, $T_l$ or $S_k$, depend on the vectors of {\em partial} residuals, which subtracts off the partial fit corresponding to the forests \textit{without} the current tree from the observed response variable: 
\begin{equation}\label{eqn:partialresidual}
  \begin{aligned}
    \res_{-l}^T & \equiv \res + a u(\X,\hat{\pi};T_l,\m_{l}^T),\quad l = 1,\dots, L, \\
    \res_{-k}^S & \equiv \res + \bvec \cdot v(\X;S_k,\m_{k}^S), \quad k = 1,\dots, K.
  \end{aligned}
\end{equation}

With these terms defined, the sequence of stochastic updates is as follows:

\begin{enumerate}
\item \textbf{Stage 1: update prognostic forest.} We first grow $L$ trees comprising the forest for the prognostic term $\mu(\x_i, \widehat{\pi}_i)$. For each of the trees ($l = 1, \dots, L$) the sequence of updates is the following:
\begin{enumerate}
  \item $T_l, \m_l^T \mid \res_{-l}^T, \sigma_0^2, \sigma_1^2, a, b_0, b_1$,  which is done compositionally as
        \begin{enumerate}
          \item $T_l \mid  \res_{-l}^T$, $\sigma_0^2$, $\sigma_1^2$
          \item $\m_l^T \mid T_l, \sigma_0^2, \sigma_1^2, a, b_0, b_1$
        \end{enumerate}
  \item $a \mid \taures, T_l$
  \item $b_0, b_1 \mid \vres, T_l$
  \item $\sigma_0^2, \sigma_1^2 \mid \res$.
\end{enumerate}

\item \textbf{Stage 2: update treatment forest.} We then grow $K$ trees comprising the forest for the treatment term $\tau(\x_i)$. The sequence of updates for each tree ($k = 1, \dots, K$) is similar:

\begin{enumerate}
  \item $S_k, \m_k^S \mid \res_{-k}^S, \sigma_0^2, \sigma_1^2, a, b_0, b_1$, which is done compositionally as
        \begin{enumerate}
          \item $S_k \mid  \res_{-k}^S$, $\sigma_0^2$, $\sigma_1^2$
          \item $\m_k^S \mid S_k, \sigma_0^2, \sigma_1^2, a, b_0, b_1$
        \end{enumerate}
  \item $a \mid \taures, S_k$
  \item $b_0, b_1 \mid \vres, S_k$
  \item $\sigma_0^2, \sigma_1^2 \mid \res$,
\end{enumerate}
\end{enumerate}

These two stages are repeated $I$ times, which we refer to as ``sweeps". Pseudocode is given in Algorithm 2. Although we use conditioning notation, note that these stochastic updates are {\em not} full conditional distributions in the usual Gibbs sampling sense. The tree-growing updates (Stage 1(a) and Stage 2(a)) are given in Algorithm 1, applied to the partial residuals defined in expression \ref{eqn:partialresidual}. Parameter updates are detailed in the next subsection.

After $I$ sweeps, the CATE estimate for individuals with features $\x$ is calculated as an average of the $(b_1-b_0)\tilde{\tau}(\x)$ samples, as if one were taking a traditional posterior mean.

\begin{algorithm}[t!]
  \small
  \caption{Accelerated Bayesian Causal Forest (XBCF)}\label{alg:XBCF}
  \begin{algorithmic}[1]
    \STATE {\bfseries Input:}  $\y, \X, L, K, I$
    \STATE  Initialize  $\res, \vres, \taures$, partial residuals $\res_{-l}^T$, $\res_{-k}^S$ and scale parameters $a, b_0, b_1, \sigma_0, \sigma_1$.
    \FOR {$\iter$ in 1 to $I$}
    \FOR {$l$ in 1 to $L$}
    \STATE  Compute partial residual $\res_{-l}^{T}$ by equation (\ref{eqn:partialresidual}).
    \STATE  Create ${\tt new\_node}$ to initialize tree $T_{l}^{\iter}$ with root node.
    \STATE  $\text{GFR}(\res_{-l}^{T}, \X,\sigma_0^2, \sigma_1^2, d = 0, T_{l}^{\iter},  {\tt new\_node})$.
    \STATE  Update leaf parameter $\m_h^{T,\iter}$ for $T_{l}^{\iter}$.
    \STATE  Update full residual $\res, \vres$ by equation (\ref{eqn:residual}).
    \STATE  Sample $a, b_0, b_1, \sigma_0, \sigma_1$ based on $\res, \vres, \taures$.
    \ENDFOR

    \FOR{$k$ in 1 to $K$}
    \STATE  Compute partial residual $\res_{-k}^{S}$ by equation (\ref{eqn:partialresidual}).
    \STATE  Create ${\tt new\_node}$ to initialize tree $S_k^{\iter}$ with root node.
    \STATE  $\text{GFR}(\res_{-k}^{S}, \X, \sigma_0^2, \sigma_1^2,  d = 0, S_{k}^{\iter}, {\tt new\_node})$.
    \STATE  Update leaf parameter $\m_k^{S,\iter}$ for $S_k^{\iter}$.
    \STATE  Update full residual $\res, \taures$ by equation (\ref{eqn:residual}).
    \STATE  Sample $a, b_0, b_1, \sigma_0, \sigma_1$ based on $\res, \vres, \taures$. \label{alg2:line:parameter}
    \ENDFOR
    \ENDFOR \\
    \STATE {\bfseries output:}  $\{\{T_l^\iter, \m_l^{T,\iter}\}_{l=1}^L, \{S_k^\iter, \m_k^{S,\iter}\}_{k=1}^K\}_{\iter = 1}^I$, $I$ posterior draws of the prognostic and treatment forests, and $\{a^{\iter}, b_0^{\iter}, b_1^{\iter}, \sigma_0^{\iter}, \sigma_1^{\iter}\}_{\iter = 1}^I$, $I$ posterior draws of other model parameters.
  \end{algorithmic}
\end{algorithm}

\subsubsection{Parameter updates}

If the no-split option is selected, or other pre-set stopping conditions are satisfied, the current node becomes a leaf node and the associated leaf parameter is updated as follows (line 8 and 16 in Algorithm \ref{alg:XBCF}). This update corresponds to a conditionally conjugate Gaussian mean update; we incorporate the control group and treatment group data sequentially to accommodate their differing variances ($\sigma^2_0$ and $\sigma^2_1$):
\begin{equation*}
  \begin{gathered}
    \nu_{n_0} = \left(\frac{1}{\nu} + \frac{n_0}{d_0^2}\right)^{-1}, \quad
    \beta_{n_0} = \frac{\bar{y}_0}{d_0^2} \nu_{n_0},  \\
  \end{gathered}
\end{equation*}
followed by
\begin{equation*}
  \begin{gathered}
    \nu_n = \left(\frac{1}{\nu_{n_0}} + \frac{n_1}{d_1^2}\right)^{-1}, \quad
    \beta_n = \left(\frac{\beta_{n_0}}{\nu_{n_0}} + \frac{\bar{y}_1}{d_1^2} \right) \nu_n,
  \end{gathered}
\end{equation*}
where $\nu$ is the prior variance over the mean, $d_0 = \frac{\sigma_0}{b_0}, d_1=\frac{\sigma_1}{b_1}$; $n_0, n_1$ are the number of individuals in control and treatment groups respectively for this leaf node, and $\bar{y}_0, \bar{y}_1$ are the corresponding partial residual means of these two groups in this leaf node. The leaf mean parameter is then sampled according to $\m \sim \N(\beta_n,\nu_n^2)$.

Model parameters $a, b_0, b_1, \sigma_0, \sigma_1$ are sampled after each  tree update, for a total of $L+K$ times per sweep. After updating trees, the model parameters are sampled based on the residual vectors in equation (\ref{eqn:residual}) -- the prognostic residual $\vres$, the treatment residual $\taures$ and the total residual $\res$ (lines 9 and 17 in Algorithm \ref{alg:XBCF}). Since the general update sequence is similar for the two stages above, we will provide an explicit update scheme of each step for only Stage 2. 

In order to update parameter $a$ we first reshape (\ref{eq:xbcfmodel}) in a regression problem where the treatment residual vector $\taures$, with each component divided by corresponding $\sigma_{z_i}$, is the response variable:
\begin{equation*}
  {\small
    \begin{bmatrix}
      \frac{y_1 - b_{z_1} \tau(x_1)}{\sigma_{z_1}} \\
      \vdots                                       \\
      \frac{y_n - b_{z_n} \tau(x_n)}{\sigma_{z_n}} \\
    \end{bmatrix}
    =
    \begin{bmatrix}
      \frac{\mu(x_1)}{\sigma_{z_1}} \\
      \vdots                        \\
      \frac{\mu(x_n)}{\sigma_{z_n}} \\
    \end{bmatrix}
    a
    +
    \begin{bmatrix}
      \frac{\epsilon_1}{\sigma_{z_1}} \\
      \vdots                          \\
      \frac{\epsilon_n}{\sigma_{z_n}} \\
    \end{bmatrix}.
  }
\end{equation*}

Then updating $a$ is essentially implemented as a two-step regression update:
\begin{equation*}
  \begin{gathered}
      \nu_{n_0} = \left(1 + \frac{\mu_0^t \mu_0}{\sigma_0^2}\right)^{-1}, \quad
      \beta_{n_0} = \frac{\taures_0^t \mu_0}{\sigma_0^2} \nu_{n_0} ; \\
      \nu_n = \left(\frac{1}{\nu_{n_0}} + \frac{\mu_1^t \mu_1}{\sigma_1^2}\right)^{-1},
      \beta_n =\left(\frac{\beta_{n_0}}{\nu_{n_0}} +\frac{\taures_1^t \mu_1}{\sigma_1^2} \right)\nu_n,
      \end{gathered}
\end{equation*}
where $\mu_0$ is a vector with elements corresponding to $\mu(\cdot)$ evaluated at rows of $\X$ for which $z_i = 0$, and similarly for $\mu_1$; $\taures_0$ is the part of residual vector $\taures$ corresponding to only individuals with $z_i = 0$, and similarly for $\taures_1$. The parameter $a$ is then sampled according to $a \sim \N(\beta_n,\nu_n^2)$.

For the scaling factors $b_0$ and $b_1$, we rearrange (\ref{eq:xbcfmodel}) in the form of a linear regression problem where the prognostic residual vector $\vres$, with each component divided by corresponding $\sigma_{z_i}$, is the response variable:
\begin{equation*}
  {\small
    \begin{bmatrix}
      \frac{y_1 - a \mu(x_1)}{\sigma_{z_1}} \\
      \vdots                                \\
      \frac{y_n - a \mu(x_n)}{\sigma_{z_n}} \\
    \end{bmatrix}
    =
    \begin{bmatrix}
      \frac{\tau(x_1)z_1}{\sigma_{z_1}} & \frac{\tau(x_1) (1-z_1)}{\sigma_{z_1}} \\
      \vdots                            & \vdots                                 \\
      \frac{\tau(x_n)z_n}{\sigma_{z_n}} & \frac{\tau(x_n) (1-z_n)}{\sigma_{z_n}} \\
    \end{bmatrix}
    \begin{bmatrix}
      b_0 \\
      b_1 \\
    \end{bmatrix}
    +
    \begin{bmatrix}
      \frac{\epsilon_1}{\sigma_{z_1}} \\
      \vdots                          \\
      \frac{\epsilon_n}{\sigma_{z_n}} \\
    \end{bmatrix},
  }
\end{equation*}
and then we update $b_0, b_1$ as regression coefficients. We first update their sampling parameters as follows:
\begin{equation*}
  {\small
    \begin{gathered}
      \nu_{n_0} = \left(\frac{1}{\frac{1}{2}} + \frac{\tau_0^t \tau_0}{\sigma_0^2}\right)^{-1}, \quad
      \beta_{n_0} = \frac{\vres_0^t \tau_0}{\sigma_0^2} \nu_{n_0} ;\\
      \nu_{n_1} = \left(\frac{1}{\frac{1}{2}} + \frac{\tau_1^t \tau_1}{\sigma_1^2}\right)^{-1}, \quad
      \beta_{n_1} = \frac{\vres_1^t \tau_1}{\sigma_1^2} \nu_{n_1},
    \end{gathered}
  }
\end{equation*}

where $\tau_0$ is a vector with elements corresponding to $\tau(\cdot)$ evaluated at rows of $\X$ for which $z_i = 0$, and similarly for $\tau_1$; $\mures_0$ is the part of residual vector $\mures$ corresponding to only individuals with $z_i=0$, and similarly for $\mures_1$. Then $b_0$ and $b_1$ are sampled as $b_0 \sim \N(\beta_{n_0},\nu_{n_0}^2), b_1 \sim \N(\beta_{n_1},\nu_{n_1}^2)$.

Lastly, updating the residual variances $\sigma_0^2$ and $\sigma_1^2$ is a conditionally conjugate inverse-Gamma update:
\begin{equation*}
  {\small
    \begin{gathered}
      \sigma_0^2 \sim \IG \left(\frac{n_0+\kappa_0}{2},\frac{2}{\res_0^t \res_0 + s_0} \right) \\
      \sigma_1^2 \sim \IG \left(\frac{n_1+\kappa_1}{2},\frac{2}{\res_1^t \res_1 + s_1} \right),
    \end{gathered}
  }
\end{equation*}
where $n_0, n_1$ are the total number of individuals fit in the control and treatment groups respectively, $\res_0, \res_1$ are the total residuals for the same corresponding groups; $\kappa_0, \kappa_1, s_0, s_1$ are hyperparameters of the inverse-Gamma prior.

\subsection{Warm-start BCF} \label{warm-start}

The simulation studies presented in Section \ref{sim_study} reveal that coverage of both BCF and XBCF often do not reach the desired nominal rate. On the one hand, complex Bayesian models do not guarantee a nominal coverage rate of credible intervals. On the other hand, very poor coverage is obviously undesirable. One contributor to under-coverage is inadequate Monte Carlo exploration of the posterior distribution, resulting in artificially narrow reported intervals. Because XBCF provides a fast approximation to the BCF posterior, initializing BCF MCMC at XBCF trees rather than roots is a promising strategy to improve the posterior exploration. Specifically, we propose the following: First, use XBCF (\textit{s} sweeps, \textit{b} burn-in) to obtain the tree draws for each of the \textit{s$-$b} sweeps after the burn-in period. Second, initialize \textit{s$-$b} BCF Markov chains at the forests obtained from XBCF. Initializing BCF on the trees obtained from XBCF substantially reduces the necessary burn-in period for the BCF MCMC algorithm. Furthermore, the separately initialized chains can be run in parallel. We call this initialization strategy warm-start BCF or ws-BCF.

In order to compare the performance and computational speed of XBCF, warm-start BCF, and the original BCF, we generated data with 50 covariates (25 continuous and 25 binary) as the input matrix and stratified treatment effects. The size of the sample is $n=5000$ and it is unbalanced on average, with approximately $\frac{2}{3}$ data points in the control group. Full details of the DGP are available in the supplement; here the time comparisons are the main interest as we expect these methods will concur on any data set given sufficient run time. 

\begin{table}[h]
\centering
{%
\begin{tabular}{c|cc|cc|cc|c}
 \toprule
     \multirow{2}{*}{Method} &\multicolumn{2}{c|}{RMSE} & \multicolumn{2}{c|}{Coverage} &  \multicolumn{2}{c|}{I.L.}& \multirow{2}{*}{Time}  \\
 & ATE & CATE & ATE & CATE & ATE & CATE \\ 
  \midrule
ws-BCF & 0.021 & 0.101 & 0.960 & 0.920 & 0.095 & 0.376 & 14 \\ 
  XBCF & 0.020 & 0.105 & 0.900 & 0.754 & 0.091 & 0.256 & 4 \\ 
  BCF(4) & 0.027 & 0.130 & 0.840 & 0.675 & 0.092 & 0.229 & 42 \\ 
  BCF(20) & 0.024 & 0.125 & 0.900 & 0.731 & 0.092 & 0.262 & 202 \\ 
   \bottomrule
\end{tabular}}
  \caption{Results of root mean squared error (RMSE), interval coverage (Coverage) and interval length (I.L.) for ATE and CATE estimators for the simulation study with 5000 datapoints and 50 covariates. The number in parenthesis for BCF indicates the number of burn-in and follow-up iterations. The column Time is running time in seconds. The results are averaged over 50 independent replications.}
  \label{tab:sim_dgp_new}
\end{table}

Results reported in Table \ref{tab:sim_dgp_new} show that warm-start BCF with default parameters (100 iterations over 40 sweeps) performs better than the original BCF MCMC in all estimands of interest, and especially improves in coverage. In general, MCMC methods need to be run for long enough in order to converge, and when we run the original BCF for a significantly larger amount of iterations (20000 after 20000 iterations of burn-in), we still see that it does not match the performance of warm-start BCF, despite taking 10 times longer.

\begin{table*}[hb!]
  \centering
  \resizebox{\textwidth}{!}{%
    \begin{tabular}{c|c|cc|cc|cc|r|cc|cc|cc|r}
      \toprule
      & \multicolumn{8}{c|}{Homogeneous Treatment} & \multicolumn{7}{c}{Heterogeneous Treatment} \\
      \midrule
      Prognostic & \multirow{2}{*}{Method} &\multicolumn{2}{c|}{RMSE} & \multicolumn{2}{c|}{Coverage} &  \multicolumn{2}{c|}{I.L.}& \multirow{2}{*}{Time} &\multicolumn{2}{c|}{RMSE} & \multicolumn{2}{c|}{Coverage} &  \multicolumn{2}{c|}{I.L.}& \multirow{2}{*}{Time} \\
      Term                      &               & ATE  & CATE & ATE  & CATE & ATE  & CATE &       & ATE  & CATE & ATE  & CATE & ATE  & CATE &       \\
      \midrule
      \multirow{8}{*}{Linear}   & ws-BCF         & 0.21 & 0.28 & 0.90 & 0.98 & 0.93 & 1.57 & 0.99  & 0.23 & 1.09 & 0.92 & 0.92 & 0.99 & 3.35 & 1.08  \\
                                & XBCF          & 0.20 & 0.24 & 0.88 & 0.94 & 0.84 & 1.13 & 0.23  & 0.23 & 1.26 & 0.86 & 0.77 & 0.86 & 2.63 & 0.24  \\
                                & BCF           & 0.23 & 0.34 & 0.88 & 0.97 & 0.92 & 1.62 & 4.64  & 0.22 & 1.14 & 0.92 & 0.81 & 0.96 & 2.93 & 4.92  \\
                                & ps-BART        & 0.26 & 0.49 & 0.87 & 0.98 & 0.99 & 2.52 & 12.44 & 0.27 & 1.21 & 0.90 & 0.93 & 1.07 & 3.67 & 12.62 \\
                                & CRF           & 0.35 & 0.54 & 0.76 & 0.86 & 1.09 & 1.58 & 0.47  & 0.40 & 1.41 & 0.78 & 0.76 & 1.23 & 2.64 & 0.44  \\
                                & BART          & 0.37 & 0.59 & 0.70 & 0.95 & 0.96 & 2.48 & 12.77 & 0.40 & 1.25 & 0.72 & 0.92 & 1.03 & 3.63 & 13.03 \\
                                & BART-$f_0f_1$ & 0.56 & 0.98 & 0.44 & 0.95 & 0.99 & 3.99 & 15.00 & 0.55 & 1.39 & 0.44 & 0.93 & 1.07 & 4.91 & 15.46 \\
                                & lm            & 0.18 & 0.31 & 0.96 & 0.99 & 0.87 & 1.73 & 2.30  & 0.22 & 0.38 & 0.92 & 0.98 & 0.97 & 1.98 & 2.14  \\
      \midrule
      \multirow{8}{*}{Nonlinear} & ws-BCF         & 0.35 & 0.44 & 0.95 & 0.99 & 1.63 & 2.56 & 0.88  & 0.38 & 1.53 & 0.90 & 0.90 & 1.59 & 4.54 & 0.97  \\
                                & XBCF          & 0.37 & 0.44 & 0.87 & 0.94 & 1.49 & 2.00 & 0.22  & 0.40 & 1.67 & 0.84 & 0.78 & 1.45 & 3.57 & 0.24  \\
                                & BCF           & 0.36 & 0.52 & 0.94 & 0.97 & 1.61 & 2.65 & 4.52  & 0.37 & 1.54 & 0.90 & 0.86 & 1.57 & 4.35 & 4.71  \\
                                & ps-BART        & 0.43 & 0.89 & 0.88 & 0.99 & 1.72 & 4.70 & 12.38 & 0.45 & 1.61 & 0.86 & 0.93 & 1.68 & 5.54 & 12.67 \\
                                & CRF           & 0.50 & 0.73 & 0.83 & 0.89 & 1.64 & 2.53 & 0.44  & 0.58 & 1.66 & 0.74 & 0.78 & 1.75 & 3.58 & 0.45  \\
                                & BART          & 0.59 & 0.97 & 0.74 & 0.97 & 1.62 & 4.44 & 12.90 & 0.58 & 1.62 & 0.70 & 0.92 & 1.58 & 5.31 & 12.90 \\
                                & BART-$f_0f_1$ & 1.38 & 2.50 & 0.14 & 0.85 & 1.70 & 7.54 & 15.02 & 1.30 & 2.65 & 0.20 & 0.86 & 1.67 & 7.86 & 15.38 \\
                                & lm            & 1.82 & 2.12 & 0.02 & 0.46 & 1.73 & 4.03 & 2.07  & 1.73 & 2.09 & 0.04 & 0.55 & 1.72 & 4.30 & 1.95  \\
      \bottomrule
    \end{tabular}}
  \caption{Results of root mean squared error (RMSE), interval coverage (Coverage) and interval length (I.L.) for ATE and CATE estimators with different combinations of treatment term and prognostic term types. Sample size is 500. The column Time is running time in seconds.}
  \label{tab:n500}
\end{table*}

\section{Simulation Study}\label{sim_study}

We reproduce the simulation study of \cite{hahn2020bayesian}, focusing on estimation of conditional average treatment effects on the basis of three metrics: average root mean square error, coverage and average interval length. The data are generated according to four different processes: the conditional expectation can be linear or nonlinear, and the treatment effect can be homogeneous or heterogeneous. The covariate vector $\x$ contains five variables, three of which are continuous, standard normal random variables, one is dichotomous, and one is unordered categorical with three levels (denoted 1,2,3). Specifically, the treatment effect is either
\[
  \tau(\x) = \left\{\begin{array}{ll}
    3            & \text{homogeneous}    \\
    1 + 2 x_2x_5 & \text{heterogeneous,}
  \end{array} \right .
\]
and the prognostic function is defined as either
\[
  \mu(\x) = \left\{\begin{array}{ll}
    1 + g(x_4) + x_1x_3      & \text{linear}     \\
    -6 + g(x_4) + 6 |x_3 -1| & \text{nonlinear,}
  \end{array} \right .
\]
where $g(1) = 2$, $g(2) = -1$ and $g(3) = -4$, and the propensity function is given by
\[
  \pi(\x_i) = 0.8 \Phi(3\mu(\x_i)/s - 0.5x_1) + 0.05 + u_i/10,
\]
where $s$ is the standard deviation of $\mu(\x)$ taken over the observed sample, with $u_i \sim \mbox{Uniform}(0,1)$. The inclusion of $\mu$ in defining the treatment probability is to induce strong confounding.

The set of methods which we use to estimate treatment effects on this data include: the two methods proposed in this paper, XBCF and warm-start BCF; the original BCF method; a naive version of BART with binary treatment assignment added as a non-distinguished covariate; ps-BART, which in addition to the treatment assignment also incorporates propensity score estimates as another covariate; BART-$f_0f_1$, which fits two separate BART models for the treatment and control groups; Causal Random Forest \citep{athey2018grf}, which also incorporates propensity score estimates \citep{grfRpackage}; and a Bayesian linear model with a horseshoe prior \cite{Carvalho2010} on the regression coefficients.

For each of the methods, we averaged the results on the three metrics over 200 independent replications. The results on a sample of $n = 500$ data points are presented in Table \ref{tab:n500}. For this simulation study, we used default recommended settings for all of the methods. Two methods, warm-start BCF and Causal Random Forest took advantage of parallelization on eight cores. 

Broadly, we recapitulate the findings of \citet{hahn2020bayesian}. Their key takeaways are that one, the propensity score is an important feature for accurate estimation of treatment effects in problems with strong confounding, and two, separate regularization of $\mu$ and $\tau$ improves estimation accuracy. Here, we highlight the differences between BCF, XBCF, and warm-start BCF.

%

\begin{itemize}
\setlength\itemsep{-0.1em}
\item XBCF provides the best CATE estimation for homogeneous treatment effect case.
\item XBCF provides the most narrow credible interval length, and often under covers compared to BCF and warm-start BCF.
\item warm-start BCF always performs better than regular BCF in CATE estimation in terms of both RMSE and coverage.
\item  Overall, warm-start BCF provides the best coverage among all three methods for both ATE and CATE.
\end{itemize}
All experiments in this paper were performed on a Linux machine with Intel(R) Core(TM) i7-8700K CPU @ 3.70GHz processor and 64GB RAM; eight cores were used for parallelization whenever it was applicable.
%

\section{Empirical demonstration}

As an empirical demonstration, we analyze data on student classroom performance in language arts class collected from two public schools in Portugal during the 2005-2006 school year \citep{cortez2008student}. This data set is publicly available at the UCI Machine Learning Repository and was used in \citet{cortez2008student} to predict students' final grades using supervised learning methods. The rich covariates in this data set make it possible to pose several questions regarding the causal impact of student's attributes on their final scores. Here, we focus on estimating the treatment effect of which school was attended, Gabriel Pereira (GP) or Mousinho da Silveira (MS). The course grade is an award on a 20-point scale.

From the original data set, which contained information on 649 students, we omit students whose final score is 0. We also restrict our analysis to those who state that they intend to pursue higher education, bringing the sample size to $n = 570$ students. 
\begin{table}[b!]
  \centering
  \begin{tabular}{rccc}
    \hline
    Method        & ATE  & CI length & Time  \\
    \hline
    ws-BCF        & 0.68 & 1.02      & 1.50  \\
    XBCF          & 0.62 & 0.91      & 0.52  \\
    BCF           & 0.67 & 1.02      & 8.02 \\
    ps-BART        & 0.67 & 0.98      & 11.07 \\
    BART          & 0.68 & 0.99      & 11.26 \\
    BART-$f_0f_1$ & 0.73 & 1.04      & 13.26 \\
    CRF           & 0.64 & 1.18      & 0.45  \\
  \end{tabular}
  \caption{ATE estimates and respective lengths of the  95\% credible/confidence intervals for the set of methods we considered.}
  \label{tab:ate}
\end{table}
We control for the following fifteen variables:
\begin{itemize}
  \setlength\itemsep{-0.1em}
  \item age: age in years at the time of the survey (numeric)
  \item address: indicator whether student lives in a city or in a rural area (binary)
  \item famrel: quality of family relationship (5 levels)
  \item famsize: indicator whether student's family has more than 3 members or not (binary)
  \item famsup: family educational support (binary)
  \item Fedu: father's education level (5 levels)
  \item Fjob: father's job (5 categories)
  \item health: student's current health status (5 levels)
  \item internet: internet access at student's home (binary)
  \item Medu: mother's education level (5 levels)
  \item Mjob: mother's job (5 categories)
  \item nursery: indicator of attending nursery school (binary)
  \item Pstatus: parent's cohabitation status (binary)
  \item reason: reason to choose this school (4 categories)
  \item sex: student's gender (binary)
\end{itemize}

%
\subsection{Treatment effect estimation}
All methods considered in the simulation study are used here as well, except for the linear model. Table \ref{tab:ate} reports point estimates and interval lengths (for 95\% credible intervals for the Bayesian methods and for the 95\% confidence interval for the Causal Random Forest method). All methods estimate the ATE to be in the range 0.6-0.8, with interval estimates lying above zero, suggesting a small positive average treatment effect.


Despite the ATE estimates broadly concurring, CATE estimates vary substantially across methods. Table \ref{tab:corr} shows the correlation matrix of CATE estimates obtained from different methods. As desired, BCF and warm-start BCF are strongly positively correlated. 
\begin{table}[t!]
  \centering
  {
    \begin{tabular}{ccccccc}
                    & CRF  & BART & BART-$f_0f_1$ & ps-BART & BCF  & XBCF \\
      \hline
      BART          & 0.65 &      &               &        &      &      \\
      BART-$f_0f_1$ & 0.63 & 0.88 &               &        &      &      \\
      ps-BART        & 0.63 & 0.87 & 0.99          &        &      &      \\
      BCF           & 0.73 & 0.62 & 0.73          & 0.71   &      &      \\
      XBCF          & 0.63 & 0.63 & 0.64          & 0.61   & 0.49 &      \\
      ws-BCF        & 0.76 & 0.73 & 0.83          & 0.82   & 0.98 & 0.57 \\
    \end{tabular}
  }
  \caption{The correlation matrix of CATE estimates obtained from different methods. }
  \label{tab:corr}
\end{table}




\begin{figure*}[b!]
  \centering
  \includegraphics[width = 1\linewidth]{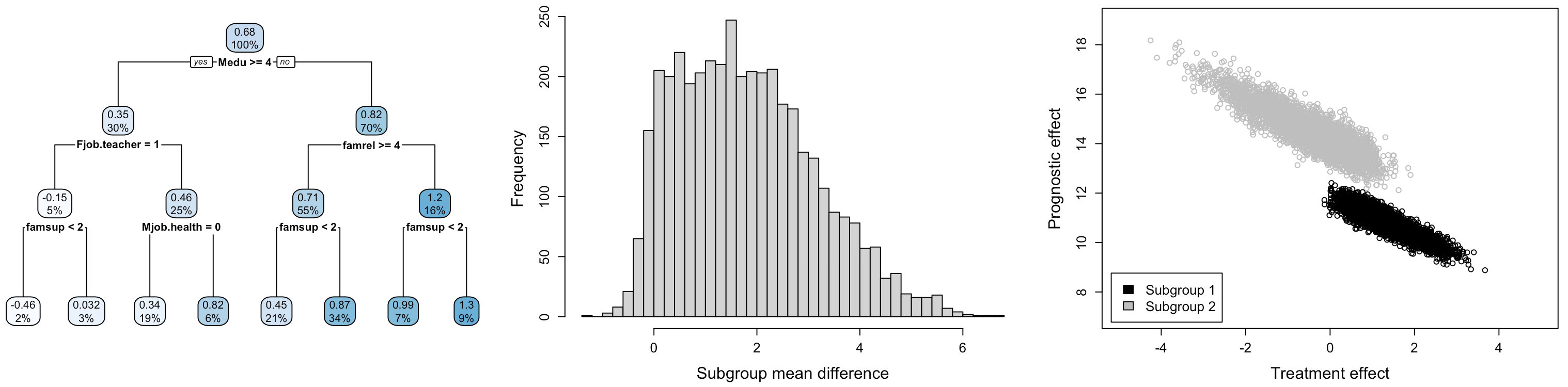}
  \caption{(Left) A single deterministic tree fit to the individual-level treatment estimates of warm-start BCF. The top number in each box is the average subgroup treatment effect, the lower number indicates the percentage of the total sample. (Middle) The histogram of difference in means of Subgroup 1 and Subgroup 2 over all posterior draws of warm-start BCF. (Right) Posterior draws of subgroup average treatment and prognostic effects for the two subgroups.}
  \label{fig:empirical}
\end{figure*}

\subsection{Subgroup analysis}

Posterior inference for subgroup average treatment effects can be obtained directly from the posterior draws sampled from warm-start BCF.


To discover subgroups of interest, we fit a regression tree to the posterior point estimates of the CATE, using the set of all covariates available from the original dataset; the resulting tree defines subgroups for which the CATE estimates differ. This should be considered a convenient form of posterior exploration and not a separate inference procedure. Posterior inferences are obtained simply as the sample average effects calculated according to each posterior draw. Of particular interest is the posterior difference between subgroup treatment effects: posterior credible intervals of this quantity allow us to determine if the difference between subgroups is statistically convincing.

The left panel in Figure \ref{fig:empirical} represents the fitted tree to posterior point estimates obtained from warm-start BCF. Subgroup 1, which benefited most from the treatment, with the subgroup ATE estimate of 1.3 points, consisted of 50 students with the following characteristics: mother doesn't have a higher education degree (Medu $<$ 4); family relationship is perceived by the student as average or lower (famrel $<$ 4); there is educational support coming from the family (famsup $\geq$ 2).

At the other end of the spectrum we have Subgroup 2, which benefited the least from the treatment, with the subgroup ATE estimate of  -0.46 points, consisting of 11 students with the following characteristics: mother has a higher education degree (Medu $\geq$ 4); father's job is teacher; there is no educational support from the family (famsup $<$ 2).

The posterior difference in subgroup ATE is shown in the middle panel of Figure \ref{fig:empirical}. The majority of the computed differences is above 0 and the 95\% posterior credible interval is $(-0.2, 4.7)$.


Although it makes intuitive sense that students whose parents have less education may stand to benefit more from better in-school instruction, the fact that those students are receiving at-home support while the children of teachers are not defied expectation. We speculate that the reason a pupil whose father is a teacher would not receive at-home support is if the student is not in need of assistance. If this were the case, it would suggest that better in-school instruction benefits students who are not already excelling; this is consistent with the estimated subgroup average prognostic effects (see right panel in Figure \ref{fig:empirical}) as well as with previous literature on educational interventions \citep{Yeager2019}.




\section{Summary}
This paper introduces a novel algorithm for fitting Bayesian causal forest models, which are increasingly popular and successful in causal inference problems with heterogeneous treatment effects. The new method makes BCF models capable of fitting larger data sets than could be fit with the previous random walk Metropolis-Hastings algorithm, which can under-explore the vast space of regression tree ensembles. We hope in the future to apply our approach to large observational health databases. Moreover, even on smaller data sets, the new algorithm provides better interval estimates of conditional average treatment effects in simulations, a property that we believe to hold for empirical analyses as well, as the warm-start BCF intervals tend to be longer. We hope that other researchers can build on these tools to consider other causal inference methods that call for regularized regression, such as instrumental variables approaches or regression discontinuity designs, et cetera.

\bibliographystyle{plainnat}
\bibliography{references}

\end{document}